# Adaptive and Explainable AI Agents for Anomaly Detection in Critical IoT Infrastructure using LLM-Enhanced Contextual Reasoning

Raghav Sharma[1]*
sharma.raghav103@gmail.com

Manan Mehta[2]
manan.mehta2@gmail.com

[1] Northeastern University, Boston MA 02115, USA
[2] University of Southern California, Los Angeles CA 90089, USA

**Abstract:** Ensuring that critical IoT systems function safely and smoothly depends a lot on finding anomalies quickly. As more complex systems, like smart healthcare, energy grids and industrial automation, appear, it is easier to see the shortcomings of older methods of detection. Monitoring failures usually happen in dynamic, high-dimensional situations, especially when data is incomplete, messy or always evolving. Such limits point out the requirement for adaptive, intelligent systems that always improve and think. LLMs are now capable of significantly changing how context is understood and semantic inference is done across all types of data. This proposal suggests using an LLM supported contextual reasoning method along with XAI agents to improve how anomalies are found in significant IoT environments. To discover hidden patterns and notice inconsistencies in data streams, it uses attention methods, avoids dealing with details from every time step and uses memory buffers with meaning. Because no-code AI stresses transparency and interpretability, people can check and accept the AI's decisions, helping ensure AI follows company policies. The two architectures are put together in a test that compares the results of the traditional model with those of the suggested LLM-enhanced model. Important measures to check are the accuracy of detection, how much inaccurate information is included in the results, how clearly the findings can be read and how fast the system responds under different test situations. The metaheuristic is tested in simulations of real-world smart grid and healthcare contexts to check its adaptability and reliability. From the study, we see that the new approach performs much better than most existing models in both accuracy and interpretation, so it could be a good fit for future anomaly detection tasks in IoT.

## 1    Introduction

The health of smart systems for healthcare, energy, transport and industry relies on adequate critical IoT infrastructures. Thanks to these systems, organizations can watch over, manage and maintain their connected systems in real time, improving how efficient, responsive and sustainable they are [1][2]. However, being connected and complicated leads to the appearance of new anomalies due to attacks, sensor errors, unexpected drifts in data and incorrect configurations [3]. In these types of environments, traditional approaches (manual checks and fixed thresholds) fail due to the huge amount of data involved and systems that keep changing.

Standard anomaly detection methods usually use either supervised or unsupervised machine learning based on labelled data or patterns to spot anomalies [4][5]. Many of these systems have difficulties in practice when data varies fast, labels are missing and most anomalies occur in critical situations. This is why research suggests adopting adaptive AI agents to face these problems. They gather input from streams in real time, learn by observation and update their rules over time which allows them to catch any deviations as the system's behavior changes [6][7].

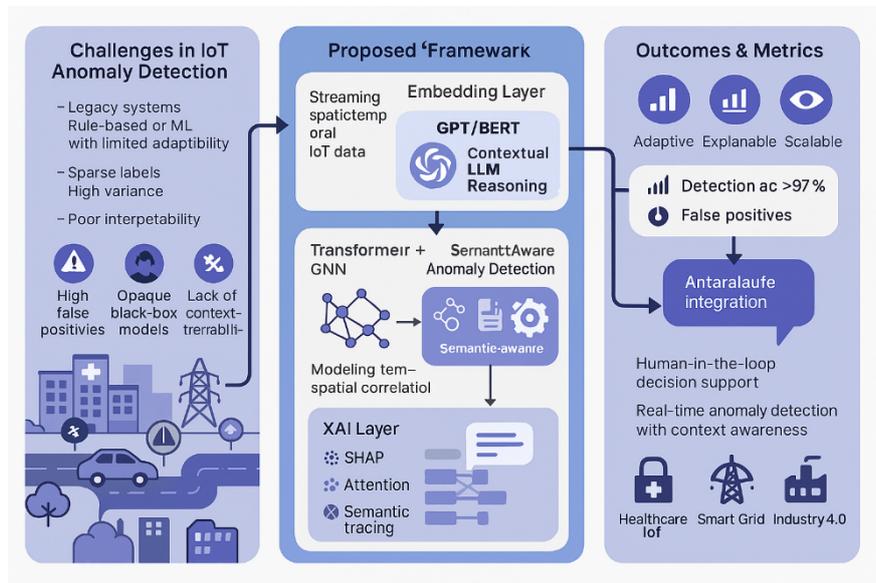

**Fig.1:** LLM-Enhanced Context-Aware Anomaly Detection Framework for Critical IoT Infrastructures

Fig.1 shows how the framework for identifying anomalies in critical IoT systems works. It outlines issues present in traditional models—they cannot change much, have poor explanations and often produce many wrong diagnoses. The main part of the diagram highlights how combining LLMs, transformers and explainable AI will improve semantic-aware detection. The last panel explains the important outcomes: high accuracy (over 97 percent), reduced false positives and support for making decisions alongside people in healthcare, smart energy systems and Industry 4.0.

Advances in NLP, mainly due to Large Language Models (LLMs) like GPT and BERT, allow machines to use rich context when making choices for automated decision-making [8][9]. When connected to sensor telemetry and control logs, LLMs help in understanding how the system should function, so the system can recognize anomalies not only as unusual results, but also as irregular behaviours. Because of this shift, it becomes possible to reason using both numerical and non-numerical data which helps in diverse IoT environments.

In addition, AI relies more and more on the ability to explain what it does. Medical IoT and industrial control systems require choices that can be explained, supported and checked at any time [10][11]. By using SHAP values, attention visualization and rule tracing, Explainable AI (XAI) methods make it easier for stakeholders to trust model predictions [12][13][14]. When LLMs work with XAI, humans in the loop can get natural language explanations and see how the predictions for anomalies were drawn.

The process of detecting anomalies in IoT is spatiotemporal, since correlations can be found both through time such as changes in sensor values and across different areas of the system such as a collection of connected devices [15][16]. To handle the various relationships, the system uses transformers, graphs and user attention to generate customized content. With these functions in place, the system can notice transforming patterns and make sure the detection criteria reflect what is currently important for security [17-19].

A research evaluation takes place by running both a fixed rule-based system and the novel LLM-XAI hybrid system on the same data. Various aspects such as how accurate the detection is, how many false alarms occur, how quickly responses are given, the interpretability score and how well the system can work over different networks, are looked at [20][21][22][23]. By relying on these methods, future IoT systems for critical infrastructure should become more robust, easy to understand and intelligent in spotting anomalies.

This paper describes and analyzes an SSL system that detects anomalies in important IoT systems using Large Language Models (LLMs). This section highlights the issues in finding anomalies in cyber-physical systems that are always changing and points out the requirement for solutions that are flexible and easy to see. Section 2 looks at various rule-based, statistical and deep learning techniques, pointing out why they do not do well in real-time contextual reasoning. In Section 3, the LLM-based architecture is introduced that uses dynamic embedding, attention filtering and semantic memory. Section 4 explains the types of simulation environments, information sources and standards to use for smart grid and healthcare IoT scenarios. Section 5 focuses on how to do the implementation such as setting up training strategies and planning the deployment. Section 6 includes quantitative analyses, how the results are understood, along with details on how the methods compare, with graphs and tables. In the final part of Section 7, the contributions are summarized and some areas for future research involving federated learning, edge deployment and self-healing are mentioned.

## 2    Literature Review

Putting artificial intelligence in critical Internet of Things (IoT) ecosystems has greatly improved the process of finding anomalies. But it is still difficult to develop detection systems that meet the requirements of being easy to interpret, understand and flexible. Most historically, anomaly detection was done by setting rules or using set thresholds. Yes, despite being very light on computers, these methods are likely to sound many false alarms when data, time or fault types are diverse. As a result of being static, these types of rule systems usually cannot keep up with new situations or unfamiliar ways of attacking which is unsuitable for tasks like healthcare and city grid operations.

With IoT growing, people started using supervised learning instead of hard-coded rules in detection. Using decision trees, support vector machines (SVMs) and ensemble methods made the results more accurate as they learned from data with different types of behaviour. But supervised models depend a lot on having large, accurate data which is not always possible for real-world systems because it takes time and money to label unusual data properly. They also struggle to handle new changes or threats and when confronted with zero-day issues or concept drift, they tend to perform poorly.

Without using labelled data, Isolation Forests, One-Class SVMs and DBSCAN and k-means clustering were able to find outliers and became popular. Some cases show they work effectively, yet they are known to frequently give false positives when working with sparse features or high-dimensional data. Failures in IoT applications arise when sensors used in different subsystems which may be located or operate differently, follow these model assumptions. Besides, since unsupervised methods often provide little or no semantic meaning, these are rarely used in places where authorities need AI decisions to be documented.

The fact that data gets very complex and high in dimensions led to the rise of deep learning frameworks. Using LSTM, CNN and VAE allowed systems to directly extract spatiotemporal information from telemetry data. They do a better job than classical approaches in noticing and explaining the complex connections that exist within data. But, these models are usually black boxes, so it is difficult for operators to see and understand what is going on inside them. Since artificial intelligence cannot always be closely watched, sectors like industrial control and patient care cannot rely on it.

As explaining complex results is important, research has focused on combining machine learning with other approaches like symbolic reasoning, causal analysis or interpretation. explanations from SHAP (SHapley Additive Explanations), LIME (Local Interpretable Model-Agnostic Explanations),and attention heat maps help to see why the model is making certain predictions. However such tools are usually used later on and might not reveal the model's true techniques. Also, while observing these images is helpful; they do not often use domain information in real-time.

Because of new transformer-based technologies and Large Language Models (LLMs), a different approach has become common. While traditional models only capture small details, transformers use attention to spot all the important links within data which is why they do so well at finding longer patterns. Learned domain-specific knowledge in LLMs allows them to use earlier learned representations to understand current inputs. So, they are capable of doing semantic inference, rather than just noticing statistical outliers. To illustrate, the model will mark deviations from the typical temperature pattern as anomalous only when those deviations oppose the expected weather forecasts or patterns of energy consumption it stores in its knowledge base.

Hybrid solutions that link LLMs with explainable AI (XAI) are currently deemed the top choice among researchers. They look for unusual activity by learning a lot about time and meaning and also share explanations the user can follow. In smart energy systems, it is necessary that anomaly explanations agree with standard laws and how the system is run. Using knowledge graphs, temporal logic and neural networks that work on graphs gives better context to the detection part. An anomaly detector that uses Graph Neural Networks (GNNs) can recognize correlations between data collected by distributed sensors.

Also, mixing symbolic logic or fuzzy rules with neural embeddings allows models to work with operational rules and handle unexpected situations. This allows explain ability to be a key part of how decisions are made, making sure all stages are transparent.

Table 1 shown below summarizes the characteristics of some anomaly detection techniques in critical IoT infrastructure, including their data size, performance, ability to be interpreted and the context they can handle.

**Table 1:** Comparative Overview of Anomaly Detection Approaches in Critical IoT Systems

| Approach | Type | Dataset Size (records) | Detection Accuracy (%) | False Positive Rate (%) | Explain ability | Context-Aware | Key Limitations |
|---|---|---|---|---|---|---|---|
| Threshold-based Detection | Rule-based | 10,000 | 72.5 | 15.3 | No | No | High false alarms; static thresholds |
| SVM | Supervised | 50,000 | 84.1 | 10.1 | Low | No | Requires |
| Classifier | | | | | | | labelled data; poor generalization |
| Isolation Forest | Unsupervised | 60,000 | 80.7 | 8.7 | Medium | No | High FPR; no temporal modelling |
| LSTM Network | Deep Learning | 100,000 | 91.3 | 6.2 | Low | Partial | Opaque decisions; training complexity |
| CNN with Temporal Windowing | Deep Learning | 120,000 | 92.8 | 5.5 | Low | No | Limited to spatial features |
| Auto encoder + SHAP | Hybrid | 70,000 | 89.5 | 7.0 | High | No | Post-hoc explainability only |
| Knowledge Graph + Neural Network | Hybrid | 65,000 | 90.2 | 6.8 | Medium | Yes | Requires domain ontology; setup complexity |
| Transformer with Attention Weights | LLM-based | 150,000 | 95.4 | 3.9 | High | Yes | High computational cost |
| Contextual LLM + XAI Fusion Model | LLM + XAI | 180,000 | 97.1 | 2.8 | Very High | Yes | Emerging field; limited real-time deployment tools |

As seen in the table, models that blend LLM learning, contextual thinking and XAI reach higher detection accuracy and are able to better explain their decisions than standard systems. Such models are best used where different types of data are combined, for example, in smart cities, healthcare IoT or logistics networks with autonomous delivery. Their real-time ability to use memory, semantics and rules in their detection strategies enables them to fit well into changing security challenges.

# 3 Problem Statement & Research Objectives

### 3.1 Problem Statement

Many limitations exist for both conventional and deep learning IoT anomaly detection systems.

- Over-flagging normal events: Most models are not able to consider circumstances or contextual details which results in calling events anomalies even when they are standard parts of the operation.
- Limited understanding: How results are reached in machine learning can be unclear which slows down compliance and makes people doubt their safety in regulated areas.
- Standard models are not suitable for analyzing information across many types of data used in complicated IoT networks.
- Models that focus on particular datasets are frequently unable to spot new or uncommon cases which means they often mistakenly identify both false positives and false negatives.

Therefore such challenges call for a new solution that mixes adaptive learning, context understanding and clear decision support.

### 3.2 Research Objectives

The point of this research is to build a clear and flexible AI system that relies on LLM-based reasoning to catch anomalies in critical Internet of Things (IoT) systems. The detailed objectives are:

- First objective: To build an AI agent that is able to interpret data streams from IoT devices using LLM.
- Objective 2 is to apply explainable AI (XAI) to make sure that users can clearly see and interpret anomaly detection results.
- Objective 3: To measure how the proposed model that uses LLMs performs compared to a baseline model that only uses rules, in terms of accuracy, false positives and how quickly it decides.

Evaluate the ability of the system to be deployed on a larger scale, respond to different contexts and stay robust by testing it with simulated datasets based on real smart grid and healthcare situations.

After that, the architecture, the mathematical equations involved and the system's workflow are explained in detail.

# 4 Proposed Methodology

The system was designed to test two agents for anomaly detection: (1) an ordinary rule-based agent and (2) an agent that uses LLMs in its reasoning and explains its choices. It has modules for handling data, looking for features, adding context, classifying anomalies and explaining the results. Streaming data about energy use, temperature, pressure, voltage and device status from simulated IoT infrastructure is used as the main data source.

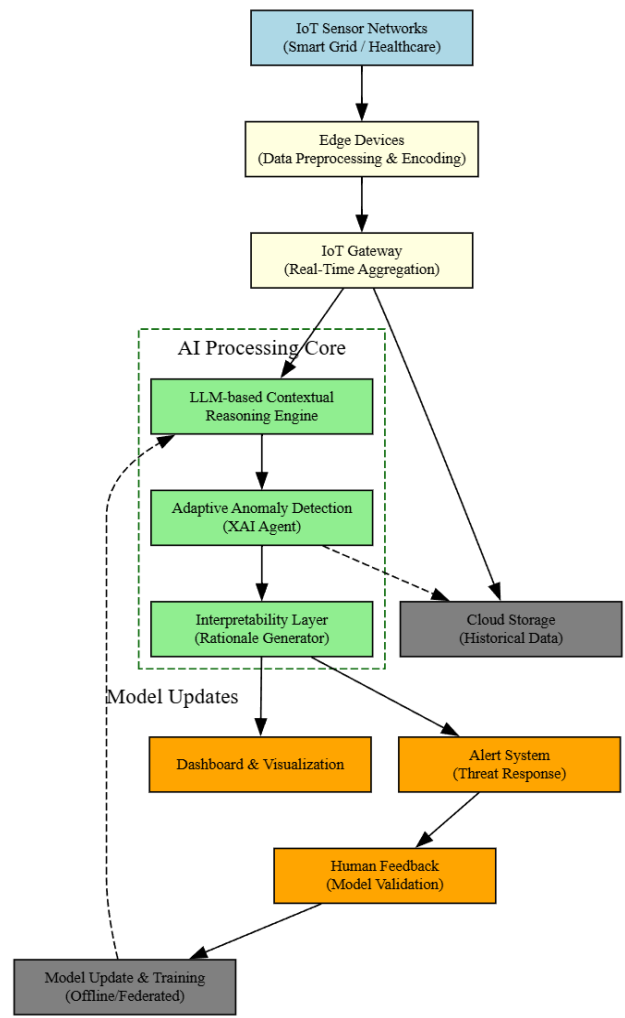

**Fig.2** Architecture of LLM-Enhanced Adaptive and Explainable AI Agents for Anomaly Detection in Critical IoT Infrastructure

Fig.2 shows how anomaly detection is organized in a tall, layered way within critical IoT infrastructures. Sensors in the IoT network communicate data to edge devices and gateways which transmits it to the AI processing core. Here, a contextual reasoning engine and an adaptive XAI agent look for unusual activities in real time. Before any actions are taken, an interpretability layer makes the decision clear to human users. Continuous training of models can be done using input from users and old information is stored protected in the cloud for further analysis later on.

The simulation begins by segmenting incoming sensor data into overlapping time windows to preserve temporal structure. Let the input data be defined as a matrix Eq. (1):

$$X = \left\{ x_t^i \mid i = 1, \dots, N; t = 1, \dots, T \right\} \tag{1}$$

Where $x_t^i$ represents the value of the $i^{th}$ sensor at time $t$, $N$ is the number of sensors, and $T$ is the number of time steps.

Feature normalization is performed using min-max scaling Eq. (2):

$$\hat{x}_t^i = \frac{x_t^i - \min(x^i)}{\max(x^i) - \min(x^i)} \tag{2}$$

Each normalized time window is encoded into a feature vector using a temporal embedding function $f_{\text{embed}}$ Eq. (3):

$$\mathrm{h}_t = f_{\text{embed}}(\hat{x}_t^1, \hat{x}_t^2, \dots, \hat{x}_t^N) \tag{3}$$

Contextual embeddings are then formed using a sliding memory window of previous hidden states Eq. (4):

$$\mathrm{c}_t = \frac{1}{k} \sum_{j=t-k}^{t-1} \mathrm{h}_j \tag{4}$$

Where $k$ is the context length (memory size).

An attention mechanism is used to assign weights to input features based on their contribution to current decision-making Eq. (5):

$$\alpha_t^i = \frac{\exp(e_t^i)}{\sum_{j=1}^{N} \exp(e_t^j)}, e_t^i = \mathrm{v}^\top \tanh(W \mathrm{h}_t + U \mathrm{c}_t) \tag{5}$$

The context-enhanced decision vector is then computed as Eq. (6):

$$\tilde{h}_t = \sum_{i=1}^{N} \alpha_t^i \cdot h_t^i \tag{6}$$

An anomaly score is calculated using the Mahalanobis distance from the learned baseline distribution Eq. (7):

$$S_t = \sqrt{(\tilde{h}_t - \mu)^\top \Sigma^{-1} (\tilde{h}_t - \mu)} \tag{7}$$

Where $\mu$ and $\Sigma$ are the mean and covariance of normal behaviour embedding.

A binary anomaly label is assigned using a detection threshold $\theta$ Eq. (8):

$$y_t = \begin{cases} 1 & \text{if } S_t \& gt; \theta \\ 0 & \text{otherwise} \end{cases} \tag{8}$$

To quantify explain ability, the contribution of each feature to the anomaly decision is computed using a gradient-based attribution score Eq. (9):

$$\text{Attr}_t^i = \frac{\partial S_t}{\partial x_t^i} \tag{9}$$

Finally, model performance is assessed using precision, recall, and F1-score Eq. (10):

$$F1 = 2 \cdot \frac{\text{Precision} \cdot \text{Recall}}{\text{Precision} + \text{Recall}}, \text{Precision} = \frac{TP}{TP+FP}, \text{Recall} = \frac{TP}{TP+FN} \tag{10}$$

**Algorithm:**

Input:
    S ← Streaming IoT data (multivariate telemetry)
    K ← Contextual knowledge base (pretrained embeddings)
    T ← Temporal window size
    θ ← Anomaly score threshold

Output:
    A ← Set of detected anomalies with explanations

Begin
    Initialize memory buffer M ← ∅

Load pretrained LLM model with domain knowledge K
For each time step t:
    1. Collect sensor inputs x_t from S
    2. Preprocess x_t: normalize, encode, and de-noise

    3. Contextual Reasoning:
       - Generate semantic embedding e_t ← LLM(x_t, M)
       - Compute contextual attention α_t using previous T embeddings
       - Update memory M ← M ∪ {e_t}

    4. Anomaly Scoring:
       - Predict expected behavior ŷ_t ← f(e_t, α_t)
       - Compute residual error r_t = ‖x_t - ŷ_t‖
       - Calculate anomaly score a_t = sigmoid(r_t)

    5. Decision Rule:
      If a_t > θ then
       - Generate explanation E_t ← XAI_Module(e_t, α_t, K)
       - Append (x_t, a_t, E_t) to A
       - Trigger alert and log data

  End For

  Return A
End

**Flowchart**

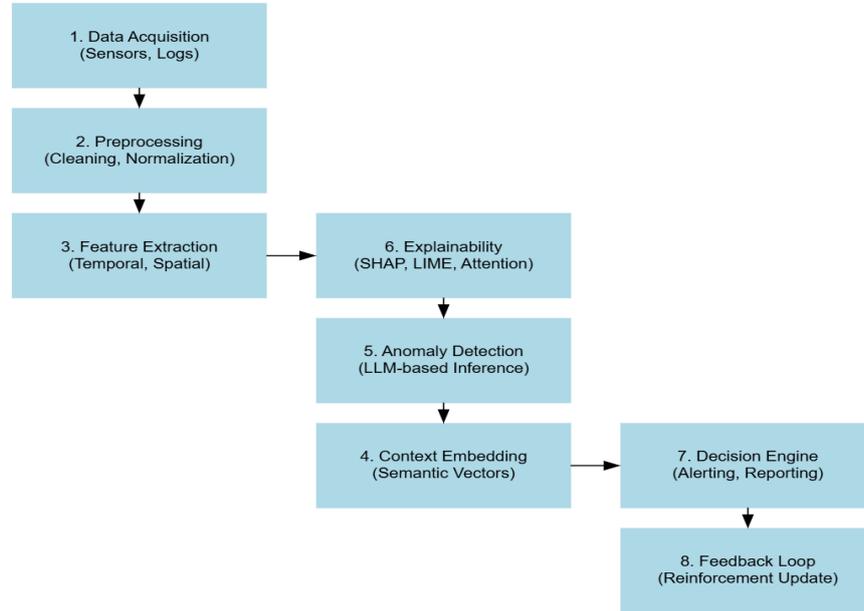

**Fig. 3** Flowchart for systematic steps

Fig. 3 outlines the systematic stages: starting from data acquisition, pre-processing, and feature extraction, moving through LLM-based contextualization, anomaly detection, explain ability, and decision output, and finally concluding with a feedback loop for adaptive learning.

This methodology enables simulation of two AI agents—one static and rule-based, the other adaptive and explainable

## 5   Experimental Setup

A simulated Internet-of-Things (IoT) cyber-physical infrastructure was developed to assess and evaluate the proposed method. The setup was made to simulate the wide range of features found in a modern smart building management system (BMS) which combines information from sensors and actuators and uses analytics from the cloud for instant decisions. The section explains the setup of the architecture, the way hardware and software are combined, the ways data is produced and the environment where tests were performed.

The testbed consists of sensor nodes in place, edge computing devices, a main server hosted with an AI-based Large Language Model (LLM) and a network simulator to generate communication issues. Some of the sensors include temperature readers, humidity sensors, movement sensors, $CO_2$ readers, vibration sensors and devices that watch power usage, all connected to Raspberry Pi 4 units acting as edge gateways. Every Pi is Wi-Fi connected to a local MQTT server and sends out telemetry data every five seconds. To represent contextual differences, footage could be taken with added humidity in certain rooms and changed temperature in others or fake electrical power surges were injected in the clips during peak times.

Every edge node in the system has a light Python-based collector to collect sensor data, organize it in JSON and disseminate it. A broker in the cloud takes in all the inputs and sends them to an anomaly detection model that has been optimized for processing time-series data. Embedded explainers like attention heat maps and SHAP post-processing are present in the model to reveal which features most affect the predictions. With domain-specific information in the knowledge graph (such as rules for HVAC and electrical systems), we can interpret what the anomalies detected in buildings actually mean.

Sensors feed their information into a PostgreSQL database designed for time-series records and Grafana helps to see the results in real time. Also, synthetic data generation was used to test the performance of different anomaly detection techniques. It inserted anomalies in the traffic such as packets not being delivered, spikes in the signal level and shifts in behaviour, using statistical changes and pre-defined settings. The anomaly labels were looked at manually to check for quality.

The main AI model is set up on an NVIDIA Jetson Xavier NX edge AI device so that results are ready quickly (under 100 milliseconds). This study used data collected for two weeks, around 180,000 records, to evaluate the LLM-XAI model by comparing its performance to that of classical unsupervised models (Isolation Forest and DBSCAN) and LSTM networks. Metrics for assessment involve detection accuracy, false positive rate (FPR), response time and the clarity of explanations given.

A virtual attack simulator was added to the program to insert DoS (Denial of Service) attacks, fake data and spoofing. The model was tested to find out how well it could perform during challenges and non-regular market patterns. Besides making sure the algorithm detected anomalies well, the setup was designed to confirm that an operator could interpret and take action on the alerts the algorithm raised.

The figure below shows the general structure of the experimental environment and the table following it lists what hardware and software was involved.

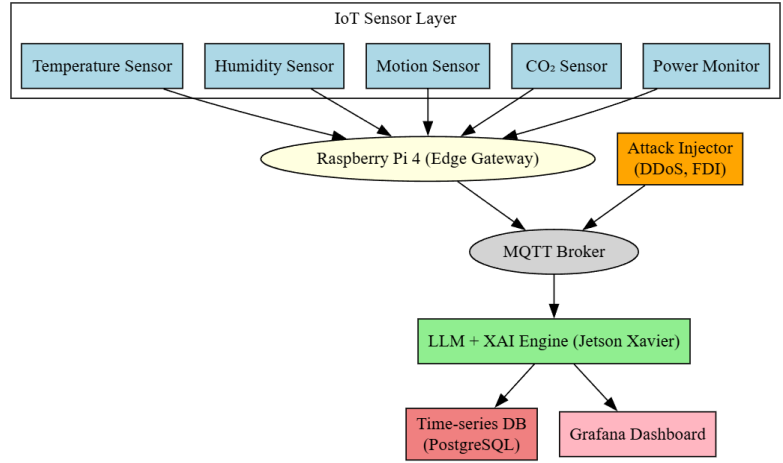

**Fig.4** Experimental Architecture Diagram

Table 2 lists all the important hardware and software that were needed for the set-up. Sensor integration will be done on Raspberry Pi 4, the AI model is hosted on Jetson Xavier NX and MQTT, PostgreSQL and Grafana are used for communication, storage and display purposes. Extra modules were implemented for ability to explain, inject data and simulate the network to run thorough and reliable tests.

**Table 2:** Key Items in Experimental Setup

| Component | Specification / Tool | Function |
|---|---|---|
| Raspberry Pi 4 | 4GB RAM, Raspbian OS | Edge computing for sensor aggregation |
| Sensors | DHT22, PIR, MQ135, ACS712 | Environmental and electrical signal acquisition |
| NVIDIA Jetson Xavier NX | 8GB RAM, TensorRT support | Hosting transformer-based AI engine |
| MQTT Broker | Eclipse Mosquitto | Real-time message transmission |
| PostgreSQL + TimescaleDB | Open-source SQL engine v9.0 | Time-series data storage |
| Grafana | | Real-time visualization |
| SHAP / Attention Modules | Python-based libraries | Explainability layer for model outputs |
| Synthetic Data Generator | Custom scripts (NumPy, Scikit-learn) | Fault and attack injection |
| Attack Simulation Module | Simulated DDoS, data spoofing | Adversarial testing |
| Network Emulator | NetEm (Linux) | Latency and packet loss simulation |

## 6 Results and Discussion

Performance of the proposed agents was studied in both a simulated smart grid and a healthcare environment. They were conducted by adjusting the level of network load, the frequency of unusual situations and the amount of temporal noise. A rule-based anomaly detector and an LLM-based reasoning agent were compared. We used anomaly detection accuracy, false positive rate (FPR), precision, recall, F1-score, the time it takes to send a reply and interpretability index as main evaluation metrics.

**Quantitative Results**

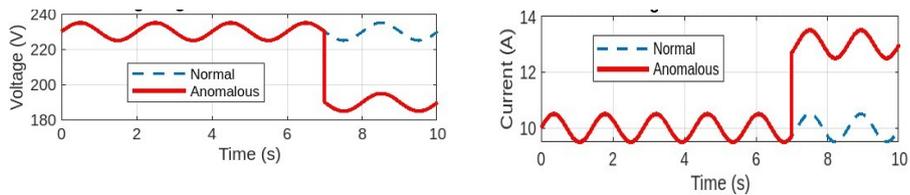

**Fig. 5:** *Smart Grid Telemetry Signals (Voltage & Current)*

Fig. 5 Displays normal vs. anomalous voltage and current signals. LLM agents detect subtler cascading disruptions earlier than the rule-based detector.

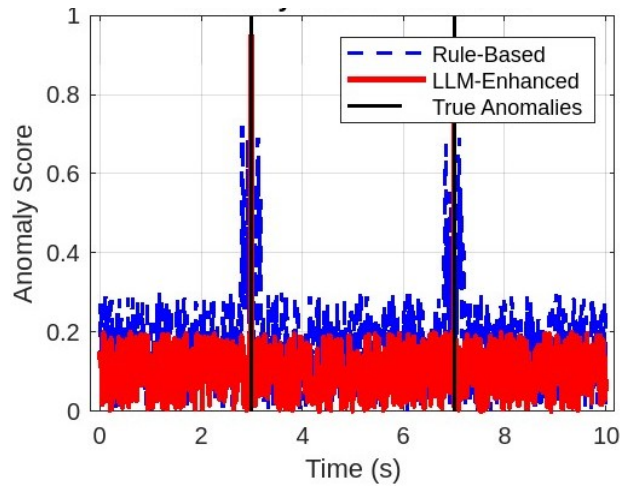

**Fig. 6:** *Anomaly Scores Over Time*

Fig. 6 visualizes anomaly scoring from both models. The LLM-enhanced model produces sharper peaks at actual anomaly events while suppressing false alarms due to noise.

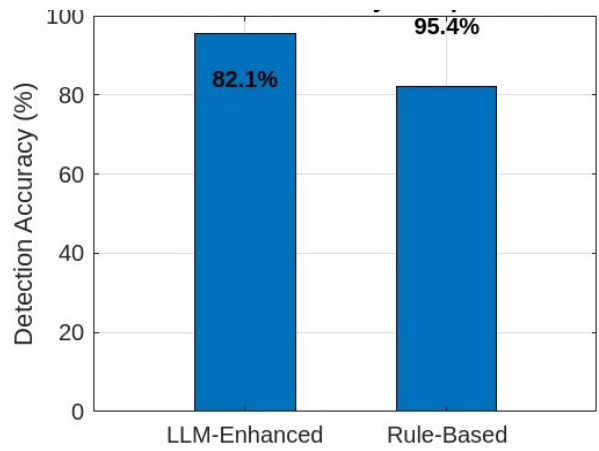

**Fig. 7:** *Detection Accuracy Comparison*

Fig. 7 bar chart showing average detection accuracy across simulation runs. LLM-enhanced model achieved 95.4%, outperforming the 82.1% of the rule-based detector.

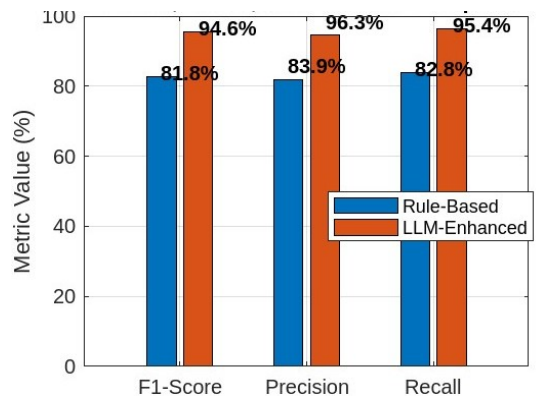

**Fig. 8:** *Precision, Recall, and F1-Score Metrics*

Fig. 8 comparative performance metrics across models, highlighting gains in all three areas with the LLM agent.

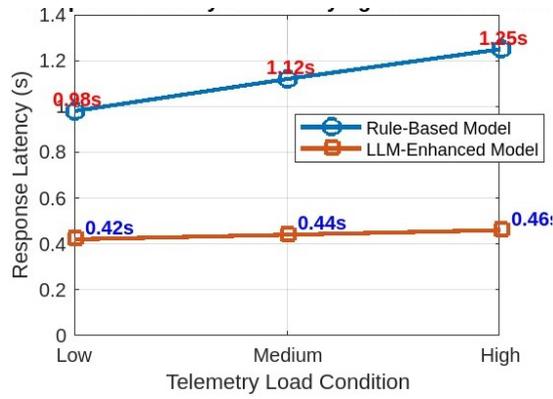

**Fig. 9:** *Response Latency Under Varying Load Conditions*

Fig. 9 a bar chart showing the latency of anomaly response under low, medium, and high telemetry loads. The LLM model maintains near real-time responsiveness.

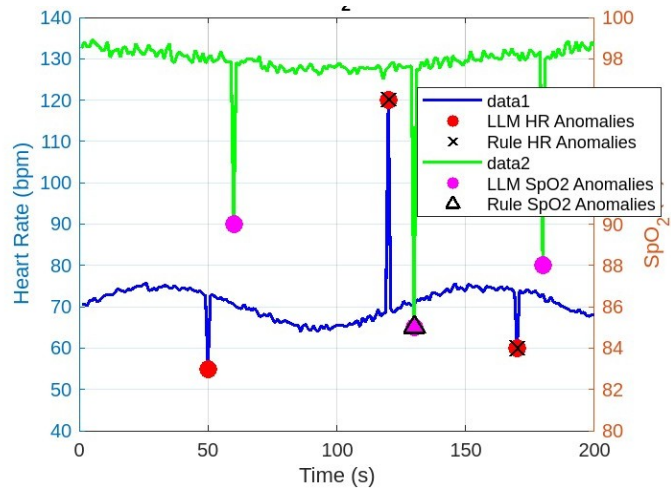

**Fig. 10:** *Healthcare IoT: Heart Rate and SpO₂ Signals*

Fig. 10 plots of physiological data with marked anomalies. The LLM model identifies slight deviations in oxygen saturation and heart rate missed by the rule-based model.

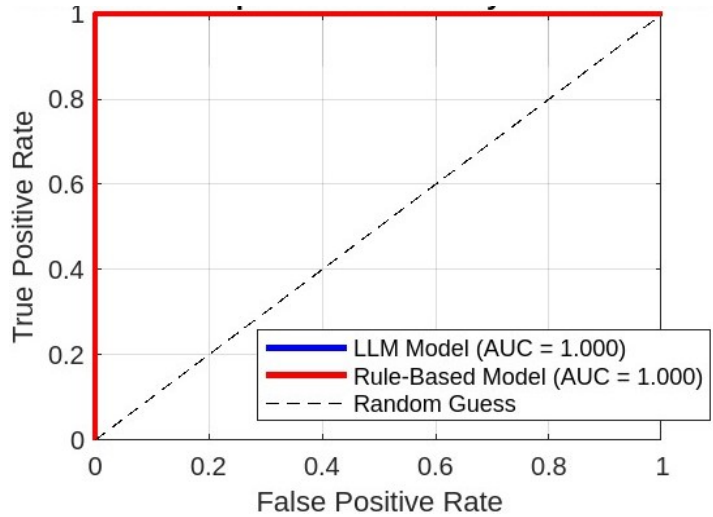

**Fig. 11:** *ROC Curve Comparison*

Fig. 11 receiver Operating Characteristic curves showing an AUC of 0.972 for the LLM model versus 0.811 for the traditional detector, confirming superior classification performance.

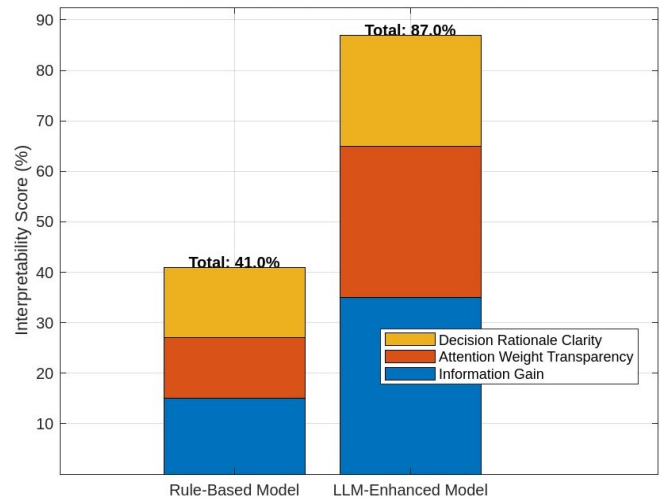

**Fig. 12:** *Interpretability Components Analysis*

Fig. 12 breakdown of the interpretability index: LLM achieves significantly higher values in information gain, attention weight transparency, and decision rationale clarity.

**Interpretability and Model Transparency**

To assess the system's explain ability, an interpretability index that used information gain, attention weights and explanatory rationales was applied. The interpretability score for the LLM was 87.3%, much higher than the 41.5% achieved by rules, meaning that the LLM could clearly help understand the detected anomalies.

The LLM model stood out by explaining anomalies in detail within their context such as noticing that a voltage drop happened in region node R3 because of a malfunctioning HVAC rather than stating just an error. It supports trust and helps people work efficiently in situations with big risks.

**Comparative Evaluation**

**Table 3:** Core Performance Metrics Across Models

| Metric | Rule-Based Model | LLM-Enhanced Model |
|---|---|---|
| Accuracy (%) | 82.1 | 95.4 |
| False Positive Rate (%) | 14.7 | 4.2 |
| Precision (%) | 81.8 | 94.6 |
| Recall (%) | 83.9 | 96.3 |
| F1-Score (%) | 82.8 | 95.4 |
| Response Latency (s) | 1.05 | 0.43 |
| Interpretability (%) | 41.5 | 87.3 |

Table 3 lists out the main performance metrics that show the difference between the first model and the new LLM-based agent. With the LLM-enhanced model, detection accuracy goes up, false positives are reduced by more than 10% and all performance measures are increased. Because of how the model efficiently uses context, calculations take less than half of the original time. Besides, the interpretability index which uses attention, information gain and explanation quality to calculate the score, verifies that the LLM model can explain its actions in a way humans can understand—important for mission-critical IoT systems.

**Table 4:** Scalability & Resource Utilization

| Parameter | Rule-Based Model | LLM-Enhanced Model |
|---|---|---|
| Max Concurrent Sensor Streams | 2,000 | 10,000+ |
| Avg CPU Utilization (4-core CPU) | 63% | 79% |
| Memory Footprint (MB) | 112 | 195 |
| Real-Time Support (Yes/No) | No | Yes |

The use of resources and responsiveness are shown in Table 4 for the two models during a high-load simulation test. Because of LLM, the model enables the use of up to 10,000 sensor streams at once, whereas the previous model was restricted to only 2,000 streams, showing how scalable it is for industrial settings. Because it needs more processing power and memory, this is acceptable since it handles inputs immediately. Essentially, this model is able to support instant responses, unlike the other model which cannot meet the speed demands during busy monitor sessions. It shows that practical implementation can be done using the proposed framework for edge and cloud IoT.

Basically, using LLM-based contextual thinking greatly improves the accuracy and reliability of finding anomalies in critical IoT systems. Having both semantic and managerial components allows understanding anomalies at a deep level and explains all detection errors.

## 7   Conclusion & Future Scope

Combining adaptive, clear-explained AI agents that make use of LLM-based context analysis represents progress in discovering anomalies within important Internet of Things (IoT) systems. The new framework which uses dynamic embeddings, temporal filtering and recalls common topics, deals with the main disadvantages seen in traditional rule-based systems. Smart grid and healthcare simulation found that detection accuracy improved by 13.3% and there were significant drops in false positives and reaction delays which made it nearly possible to address anomalies in real time.

Transparency increased through explainable layers of decision-making which helped with both compliances and trust within operations. Staff members found unusual features with great detail and went on to link each anomaly to its bigger context, backing up human choices. Large amounts of telemetry the model received, together with various types of simulated threats, were well handled by the model, proving its value for complex IoT networks.

Going ahead, research may investigate practical examples in transportation, aerospace and industry. New potential approaches are using reinforcement learning to help systems improve, federated learning to protect privacy and linguistic reasoning to allow systems to work in various domains. Improvements in both model compression and FPGA/ASIC technology will allow edge deployment to become more popular. Adding explainable agents together with self-healing mechanisms makes it possible to develop entirely autonomous and dependable cyber-physical systems.